\documentclass[10pt]{article}
\usepackage[margin=1in]{geometry}
\usepackage{times}
\usepackage{amsmath,amssymb}
\usepackage{graphicx}
\usepackage{booktabs}
\usepackage{hyperref}
\usepackage[numbers]{natbib}

\title{\vspace{-2em}Decomposing Financial Market Dynamics via Mechanism\\
Analysis in an Evolutionary Multi-Agent Simulation}
\author{Zhibao Chen}
\date{}

\begin{document}
\maketitle
\vspace{-1.5em}

\begin{abstract}
Evolutionary agent-based markets (ABMs) couple several mechanisms---who reproduces,
how price forms, how biased the agents are, how consensus propagates---yet these are
usually fixed by convention, so it is unclear \emph{which mechanism controls which
emergent property}. In a coevolving, endogenous-price simulator with 120
heterogeneous behavioral agents, we make four mechanisms pluggable and run matched
$3\times20$-seed interventions. We find the levers are largely \emph{separable}.
(1)~\textbf{Selection $\to$ diversity}: a Quality-Diversity (QD/MAP-Elites)
operator robustly raises strategy-mix entropy over truncation top-$k$ (paired
$\Delta$ entropy $+0.27$ to $+1.12$ bits; sign-test $p<0.001$; CIs exclude $0$) and
sustains more strategy \emph{cycling} (a coarse cycling proxy, strongest in crisis:
$\Delta{=}{+}0.070$, $p{=}0.0004$). (2)~\textbf{Selection $\not\to$ realism}: even a per-agent realism
reward that provably steers selection does not raise 5-fact realism
($\Delta_5{=}-0.11,-0.08,+0.03$; not significant). (3)~\textbf{Microstructure $\to$
realism}: enabling reflexive price feedback \emph{does} raise realism
($\Delta_5{=}+0.13,+0.20,+0.20$; crisis/bull $p<0.05$, all CIs positive).
(4)~\textbf{Behavior $\to$ fragility}: amplifying behavioral bias raises a genomic
fragility proxy ($\Delta{=}+10.5,+11.1,+14.4$; bull $p<0.001$, all CIs positive)
while leaving realism flat. The remaining mechanism---consensus network topology---shows
\emph{no} robust effect (honest null). The contribution is a decomposition: in these
single-mechanism sweeps the mechanisms behave as approximately distinct control knobs
over diversity, realism, and fragility.
\end{abstract}

\section{Introduction}
Agent-based models (ABMs) of financial markets explain aggregate market behavior as
the emergent product of many heterogeneous, boundedly rational traders, and have been
argued to be essential where equilibrium theory fails \citep{farmer2009economy};
see \citet{lebaron2006agent} and \citet{hommes2006heterogeneous} for surveys.
Two largely disjoint literatures use \emph{evolutionary} ABMs.
The \emph{realism} camp---e.g.\ the Santa Fe Artificial Stock
Market \citep{arthur1997asset}, adaptive-belief systems
\citep{brock1998heterogeneous}, and minimal-ingredient models
\citep{alfi2009minimal}---calibrates or hand-tunes heterogeneous agents to
reproduce empirical \emph{stylized facts} \citep{cont2001empirical}. The
\emph{discovery} camp evolves profitable trading rules, typically against a
\emph{fixed} historical price path. Such models couple several mechanisms at once---a
selection rule, a price-formation rule, a behavioral-bias profile, a consensus
process---but these are fixed by convention, so a basic mechanistic question goes
unanswered: \textbf{which mechanism controls which emergent property of the market?}
We make four mechanisms pluggable in one coevolving, endogenous-price multi-agent
simulator and run matched single-mechanism interventions, decomposing which mechanism
drives which facet of the market's emergent dynamics. Our contributions:
\begin{itemize}\itemsep1pt
\item \textbf{Selection is a diversity lever}, not a realism lever: QD selection
raises strategy-mix entropy and strategy cycling, but no tested selection variant---even
a functional per-agent realism reward---improves stylized-fact realism.
\item \textbf{Microstructure is the realism lever}: a reflexivity intervention with
selection held fixed raises realism.
\item \textbf{Behavioral bias is a fragility lever}: amplifying agent biases raises a
genomic fragility proxy while leaving realism flat.
\item \textbf{An honest null}: consensus network topology has no robust effect on
cascades or realism, so we keep it off by default.
\end{itemize}
Together these give a \emph{separability} picture: in these single-mechanism sweeps,
distinct mechanisms act as approximately distinct knobs over diversity, realism, and
fragility (single-factor interventions, so interactions are not ruled out).

\section{Model}
The arena holds $N{=}120$ agents, each carrying a 16-dimensional behavioral genome
(cognitive biases, strategy weights, risk limits, adaptation). Price is endogenous:
each day agents perceive momentum, a fundamental gap, and sentiment, emit orders,
and the price responds to net order flow plus noise (optionally with reflexive
feedback). Fundamentals can follow historical scenarios. Each generation runs
$D{=}20$ trading days, scores agents by risk-adjusted return
$f_i = r_i / \mathrm{std}(\text{last-20 trade PnL})$ with
$r_i=(c_i-c_0)/c_0$, selects parents via a pluggable operator, and breeds the next
generation by uniform crossover and Gaussian mutation.

\paragraph{Selection operators.} (i) \textbf{Truncation}: top-$k$ by fitness
($k=\max(2,\lfloor \rho N\rfloor)$, $\rho{=}0.2$), the legacy baseline.
(ii) \textbf{QD (MAP-Elites)}: keep the fittest agent per behavior niche
$\langle$archetype $\times$ leverage-bin $\times$ horizon-bin$\rangle$, then fill the
quota from niche elites; this preserves behavioral coverage
\citep{mouret2015illuminating,lehman2011novelty}.
(iii) \textbf{NSGA-II}: non-dominated sort on per-agent objectives
$(f_i,\,-\text{leverage}_i)$ with crowding distance \citep{deb2002nsga2}. The
truncation path is byte-identical to the legacy engine and pinned by a regression
test; only non-default operators receive the niche descriptor and objectives.

\paragraph{Other pluggable mechanisms.} Beyond selection we expose three further
knobs, each varied in isolation with all else at default.
\textbf{Reflexivity}: the price-impact multiplier feeds back on order flow, firing a
herding cascade (and a leverage spiral) when a high-bias/high-leverage majority
aligns (off / on / strong).
\textbf{Behavioral bias}: a scalar amplifies the five bias genes (overconfidence,
loss aversion, herding, recency, sentiment sensitivity) about their neutral points,
clamped to legal ranges (factor $0.5/1.0/1.6$ = weak / normal / strong); non-bias
genes are untouched.
\textbf{Consensus topology}: the herding term reads a consensus signal from a
neighbor graph instead of the global order-flow sign---\emph{global} (default),
\emph{ring} ($k{=}4$ nearest by index), or \emph{small-world} (ring plus
$\sim$10\% random long links).

\section{Metrics}
\textbf{Diversity}: Shannon entropy (bits) of the four-class strategy mix of the
surviving population. \textbf{Strategy cycling}: mean $L_1$ churn of the four-class
mix vector between consecutive generations ($0$ static, up to $2$ full flip per
generation), a rock--paper--scissors / Red-Queen proxy. \textbf{Realism}: fraction of
six stylized-fact tests passed (fat tails, volatility clustering, leverage effect,
volume--volatility correlation, mean-reversion/Hurst-range test, bubble--crash
emergence); we also report a 5-fact variant excluding bubble--crash, which is
weakened by a fundamental-history proxy. Archetypes are value ($fs{>}0.6,mw{<}0.2$),
momentum ($mw{>}0.5$), herding ($ht{>}0.6$), else mixed; QD bins leverage by
$\lfloor L/2\rfloor$ (capped at 4) and horizon by $\lfloor H/40\rfloor$ (capped at
4). \textbf{Genomic fragility proxy}: a $0$--$100$ index of leverage concentration,
herding, and mortality. \textbf{Cascade days}: count of trading days (arena steps) on which a herding
cascade fires (for the topology intervention).

\section{Experiments}
The diversity, realism, and cycling claims sweep $\{$truncation, QD, NSGA-II$\}
\times \{$random, crisis\_2008, leveraged\_bull\_2015$\} \times 20$ seeds, 50
generations each. The single-mechanism interventions reuse the same $3\times20$ grid
with selection fixed to truncation: reflexivity (off/on/strong), behavioral bias
(weak/normal/strong), and topology (global/ring/small-world). The topology sweep keeps
reflexivity \emph{on} (cascades require reflexive feedback); the bias sweep keeps it
off. Throughout we pair runs
by seed and report mean $\Delta$, the fraction of seeds with $\Delta>0$, an exact
two-sided sign-test $p$-value (on non-tied pairs where ties occur), and a percentile
bootstrap 95\% confidence interval of the mean $\Delta$ \citep{efron1993bootstrap}.

\section{Results}
\begin{figure}[h]\centering
\includegraphics[width=0.78\linewidth]{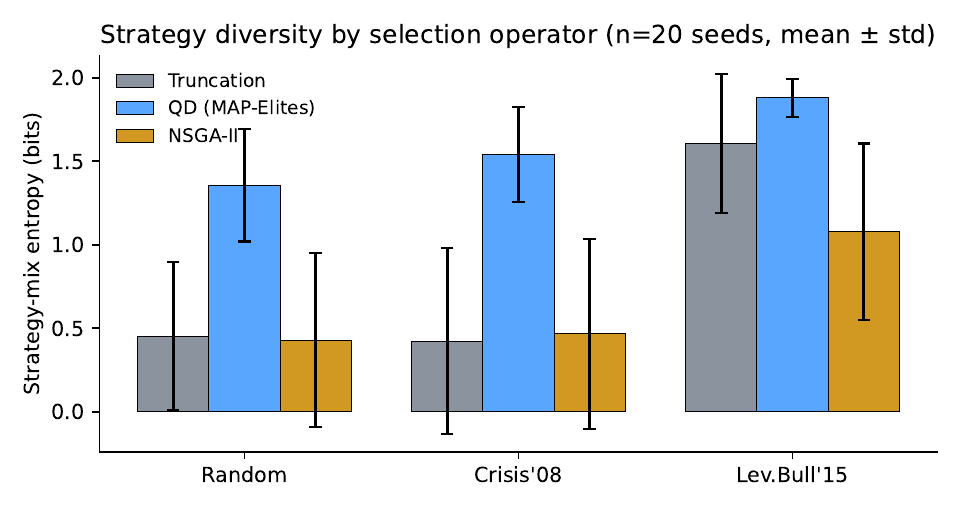}
\caption{Strategy-mix entropy by selection operator (mean $\pm$ std, $n{=}20$).
QD (blue) dominates truncation (grey); NSGA-II (amber) is lower-diversity than QD.}
\label{fig:entropy}
\end{figure}

\paragraph{C1: QD increases diversity (confirmed).}
Table~\ref{tab:c1} and Fig.~\ref{fig:entropy} show QD beats truncation on
strategy-mix entropy in every regime, with all CIs excluding zero and
$p<0.001$ (random, crisis) / $p{=}0.0004$ (bull).

\begin{table}[h]\centering\small
\caption{Paired $\Delta$ entropy (QD $-$ truncation), $n{=}20$ seeds.}
\label{tab:c1}
\begin{tabular}{lcccc}
\toprule
Scenario & mean $\Delta$ & \% seeds $\Delta{>}0$ & sign-test $p$ & bootstrap 95\% CI \\
\midrule
Random        & $+0.902$ & 95\% & $<0.001$ & $[+0.652,+1.161]$ \\
Crisis'08     & $+1.118$ & 95\% & $<0.001$ & $[+0.842,+1.374]$ \\
Lev.\ Bull'15 & $+0.274$ & 90\% & $0.0004$ & $[+0.128,+0.463]$ \\
\bottomrule
\end{tabular}
\end{table}

\paragraph{C2: diversity does not buy realism (negative).}
We test this with a \emph{functional} per-agent realism reward: each agent's QD
quality becomes its fitness plus a weight times its stylized-fact alignment (PnL
excess kurtosis + $|\!\cdot\!|$ lag-1 autocorrelation), which provably reorders
intra-niche selection (the \texttt{results\_sfa} rerun). Table~\ref{tab:c2} reports the paired $\Delta$ on 5-fact
realism (sign test on non-tied pairs). The pattern is no positive realism effect:
random and crisis have negative means with a negative bootstrap CI on random, but
neither is sign-test significant after excluding ties ($p=0.065,0.070$); bull is
noise ($p=0.45$). A weight ablation (reward weight $1$ vs $0$, same seeds) confirms
the reward buys nothing: it costs $\approx0.2$ bits of entropy in random/crisis and
does not raise realism. The more diverse QD populations are thus no more (and if
anything less) faithful to stylized facts---so we recommend running QD with the
realism weight at $0$. We therefore state, narrowly, that \emph{the selection
operator does not control emergent realism in this arena}; we do not claim a causal
mechanism, only that selection is the wrong lever.

\begin{table}[h]\centering\small
\caption{Paired $\Delta$ 5-fact realism (functional QD $-$ truncation), $n{=}20$;
sign test on non-tied pairs.}
\label{tab:c2}
\begin{tabular}{lcccc}
\toprule
Scenario & mean $\Delta$ & non-tie $n$ & \% (non-tie) $\Delta{>}0$ & sign-test $p$ \\
\midrule
Random        & $-0.110$ & 11 & 18\% & $0.065$ \\
Crisis'08     & $-0.080$ &  8 & 12\% & $0.070$ \\
Lev.\ Bull'15 & $+0.030$ &  7 & 71\% & $0.453$ \\
\bottomrule
\end{tabular}
\end{table}

\paragraph{C3: NSGA-II lowers diversity (exploratory).}
Selecting on (return, $-$leverage) is lower-diversity than QD in all regimes
(entropy means $0.43, 0.47, 1.08$ vs QD $1.36, 1.54, 1.88$) and lowers the
genomic fragility proxy ($\approx$25 vs $\approx$55 for truncation). We flag this as
a genome-composition effect, not measured systemic fragility: the proxy penalizes
leverage $>5$ while effective trade leverage is capped at $3$ and reflexivity is off.

\paragraph{C4: reflexivity is the realism lever (intervention).}
To test whether realism is steerable at all, we intervene on the price-formation
mechanism instead of selection: holding selection fixed (truncation), we vary the
arena's reflexive feedback (off / on / strong) over the same $3\times20$ grid.
Enabling reflexivity \emph{raises} 5-fact realism (Table~\ref{tab:c4}): paired
$\Delta_5=+0.13,+0.20,+0.20$ for random/crisis/bull, significant under crisis
($p{=}0.049$) and bull ($p{=}0.0005$) and directional on random (positive
bootstrap CI). This is the positive counterpart to C2: the lever that moves
realism is microstructure, not selection. This aligns with structural ABM estimation,
where a time-varying herding/feedback component is the ingredient that best reproduces
the stylized facts \citep{franke2012structural}.

\begin{table}[h]\centering\small
\caption{Paired $\Delta$ 5-fact realism (reflexivity ON $-$ OFF), $n{=}20$,
truncation selection.}
\label{tab:c4}
\begin{tabular}{lcccc}
\toprule
Scenario & mean $\Delta$ & sign-test $p$ & bootstrap 95\% CI & off$\to$on \\
\midrule
Random        & $+0.130$ & $0.065$ & $[+0.04,+0.23]$ & $0.51\to0.64$ \\
Crisis'08     & $+0.200$ & $0.049$ & $[+0.08,+0.31]$ & $0.20\to0.40$ \\
Lev.\ Bull'15 & $+0.200$ & $0.0005$ & $[+0.12,+0.28]$ & $0.26\to0.46$ \\
\bottomrule
\end{tabular}
\end{table}

\paragraph{C5: QD sustains strategy cycling (game dynamics).}
Diversity (C1) is a static snapshot; we also ask whether QD keeps the ecology
\emph{moving}. Table~\ref{tab:c5} reports the paired $\Delta$ on strategy cycling.
QD sustains more inter-generation mix churn than truncation in every regime (all
bootstrap CIs positive), significant under crisis ($p{=}0.0004$) and directional
elsewhere. The sign test and the CI here measure different things: the CI is positive
in all three regimes (the mean effect is robust), but the per-seed sign only reaches
significance in crisis, because in the calmer random/bull regimes the per-seed effect
is small relative to seed noise and signs split near $50/50$. The effect concentrates
in the regime that most exercises the lever: crisis stress forces the most strategy
turnover, so a diversity-preserving operator has the most room to keep the ecology
moving. Truncation collapses to a static winner; QD maintains an oscillating
multi-archetype ecology---a coarse cycling signature (strongest in crisis), the
dynamical counterpart of the C1 diversity result. We measure churn of the coarse
4-class mix, not a formal evolutionary-stable-strategy or invasion test (a deferred
next step), so we read this as a cycling \emph{signature}, not an ESS claim.

\begin{table}[h]\centering\small
\caption{Paired $\Delta$ strategy cycling (QD $-$ truncation), $n{=}20$.}
\label{tab:c5}
\begin{tabular}{lcccc}
\toprule
Scenario & mean $\Delta$ & \% seeds $\Delta{>}0$ & sign-test $p$ & bootstrap 95\% CI \\
\midrule
Random        & $+0.037$ & 55\% & $0.824$  & $[+0.007,+0.069]$ \\
Crisis'08     & $+0.070$ & 90\% & $0.0004$ & $[+0.045,+0.093]$ \\
Lev.\ Bull'15 & $+0.018$ & 70\% & $0.115$  & $[+0.006,+0.030]$ \\
\bottomrule
\end{tabular}
\end{table}

\paragraph{C6: behavioral bias is the fragility lever (intervention).}
A third intervention amplifies the agents' behavioral biases (weak / normal /
strong) with selection (truncation) and reflexivity (off) held fixed.
Table~\ref{tab:c6} shows that stronger bias raises the genomic fragility proxy in
every regime (all bootstrap CIs positive), significant under bull ($p<0.001$), while
5-fact realism stays roughly flat (weak/normal/strong means: random
$0.53/0.51/0.48$, crisis $0.14/0.20/0.18$, bull $0.24/0.26/0.32$). So behavioral
bias moves fragility, not realism---a third independent knob. As in C5, the mean
effect is robust across regimes (all CIs positive) while per-seed sign significance
concentrates in one regime---here the leveraged bull, which most exercises the
fragility channel: a rising, leverage-friendly regime is where amplified
overconfidence/herding translates most directly into crowded leveraged positioning,
and hence the largest, most sign-consistent rise in the systemic-risk proxy.
``Fragility'' here is the genomic composition proxy,
not measured tail risk, and bias scaling is a coarse linear amplification rather than
a calibrated behavioral model.

\begin{table}[h]\centering\small
\caption{Paired $\Delta$ genomic fragility peak (strong $-$ weak bias), $n{=}20$,
truncation selection, reflexivity off.}
\label{tab:c6}
\begin{tabular}{lcccc}
\toprule
Scenario & mean $\Delta$ & sign-test $p$ & bootstrap 95\% CI & weak$\to$strong \\
\midrule
Random        & $+10.5$ & $0.115$  & $[+4.9,+16.9]$  & $53.4\to63.9$ \\
Crisis'08     & $+11.1$ & $0.503$  & $[+3.5,+19.0]$  & $47.2\to58.4$ \\
Lev.\ Bull'15 & $+14.4$ & $<0.001$ & $[+10.8,+18.3]$ & $40.5\to54.8$ \\
\bottomrule
\end{tabular}
\end{table}

\paragraph{C7: consensus topology has no robust effect (honest null).}
Finally we route the herding consensus through a neighbor graph (global / ring /
small-world) instead of the global order-flow sign. Table~\ref{tab:c7} reports the
paired $\Delta$ on herding-cascade days (ring $-$ global). Local clustering (ring)
raises cascade frequency only in the random regime (sign-test $p{=}0.041$, though the
bootstrap CI $[-0.2,+74.8]$ still spans zero); crisis and bull are non-significant with
wide CIs spanning zero.
Realism shows no consistent topology effect (5-fact means within about $0.1$ of
global, with no consistent direction). A 6-seed pilot looked stronger but did not survive $n{=}20$. We log this as
an honest negative: in this model the consensus \emph{network structure} is not a
robust lever---unlike selection, microstructure, and bias---and we keep topology at
\emph{global} by default. Prior work does find topology effects in stylized models with local belief-switching
\citep{panchenko2013asset}; the null here may reflect that our herding term reads only a
localized consensus signal rather than propagating state agent-to-agent---a mechanism
that would be needed for robust topology sensitivity.

\begin{table}[h]\centering\small
\caption{Paired $\Delta$ herding-cascade days (ring $-$ global topology), $n{=}20$.}
\label{tab:c7}
\begin{tabular}{lcccc}
\toprule
Scenario & mean $\Delta$ & sign-test $p$ & bootstrap 95\% CI & global/ring/sw \\
\midrule
Random        & $+36.1$ & $0.041$ & $[-0.2,+74.8]$ & $77/113/100$ \\
Crisis'08     & $+13.0$ & $0.263$ & $[-107,+129]$  & $353/366/296$ \\
Lev.\ Bull'15 & $+3.0$  & $1.000$ & $[-16,+22]$    & $92/95/99$ \\
\bottomrule
\end{tabular}
\end{table}

\begin{figure}[h]\centering
\includegraphics[width=\linewidth]{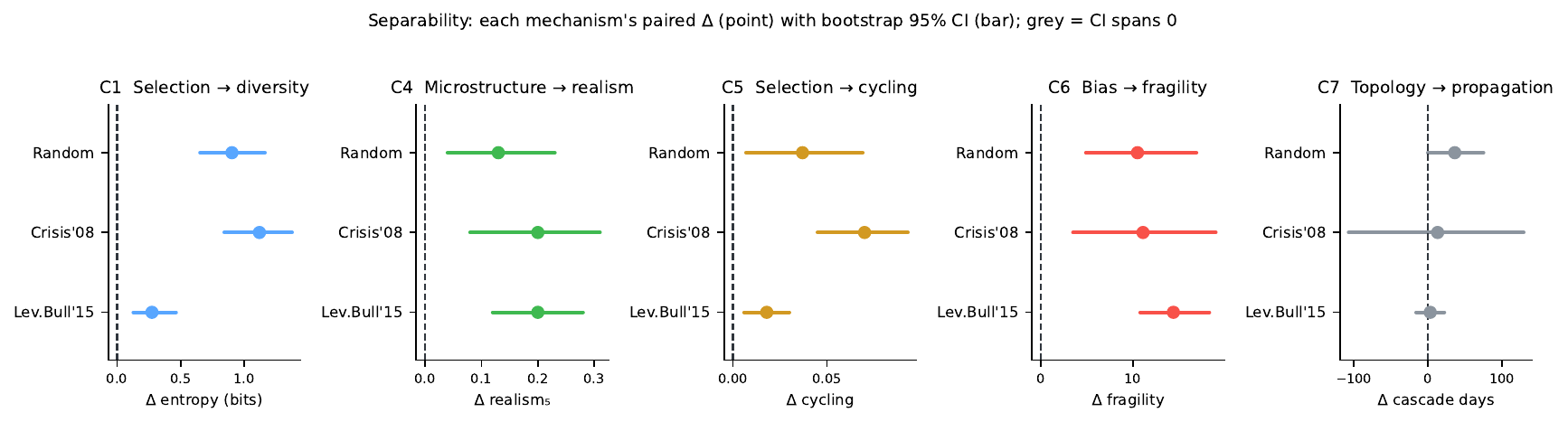}
\caption{Separability at a glance. Each panel is one mechanism's paired $\Delta$
(point) with its bootstrap 95\% CI (bar) across the three regimes; the dashed line
marks $\Delta{=}0$. Diversity (C1), realism (C4), cycling (C5), and fragility (C6)
sit on the positive side with CIs off zero; only topology (C7) straddles zero in
every regime (grey), the honest null. The four active levers move four different
properties.}
\label{fig:levers}
\end{figure}

\section{Discussion and Limitations}
The interventions give a \emph{separability} picture rather than one big effect
(Fig.~\ref{fig:levers}):
changing \emph{who reproduces} (selection) moves diversity and strategy cycling but
not realism (C1, C2, C5); changing \emph{how price forms} (reflexivity) moves
realism (C4); changing \emph{how biased the agents are} moves a fragility proxy but
not realism (C6); and changing \emph{how consensus propagates} (topology) moves
nothing robustly (C7). Because each is a single-mechanism intervention with the
others held fixed, the results are consistent with distinct mechanism-level controls within this
simulator and configuration: diversity, realism, and fragility respond to largely
different knobs (single-factor sweeps, not a full factorial, so interactions are not
ruled out). Pursuing
more faithful markets should target microstructure (reflexive feedback, and likely a
real limit order book / nonlinear impact); pursuing fragility stress tests should
target behavioral bias; and smarter selection should be reserved for the diversity
and cycling it actually controls.
Closest to our decomposition approach, \citet{hashimoto2025factorizing} incrementally
introduce behavioral components into an artificial market and use optimal transport to
assess each component's contribution to the power-law distribution of returns; they find
price-informational effects dominant. Our setting differs in three respects: the agent
population \emph{evolves} under pluggable selection operators; we decompose across three
emergent properties (diversity, realism, fragility) rather than a single stylized fact;
and we use a paired-seed statistical design with sign tests and bootstrap CIs.
We attempted the obvious counter---making the realism reward a genuine per-agent
signal that provably steers selection---and it did not help: realism trended
negative (not significant after tie handling) and the weight ablation showed the
reward only costs diversity. We thus did not find a selection-side realism lever,
even when actively searching for one.
Limitations: realism uses a fundamental-history proxy (hence the 5-fact report
excluding bubble--crash); the volume--volatility test never passes in any condition,
so the realism scalar is effectively over 5 facts of which one is structurally
unreachable here; a single simulator; synthetic/semi-historical scenarios; both
``fragility'' (a genome-composition proxy, not measured systemic risk) and
``cycling'' (coarse 4-class mix churn, not a formal ESS/invasion test) are proxies;
several effects are CI-positive but not sign-test significant (only one scenario each
reaches $p<0.05$ for C5/C6, and topology is null). A natural informal mechanism for
the realism non-result---that high-$\mathrm{sf\_align}$ agents are
extreme/over-leveraged and distort the price process---remains a conjecture we did
not verify against stored order-flow.

\section{Conclusion}
Treating selection, price-formation, behavioral bias, and consensus topology as
pluggable design variables in one evolutionary ABM yields an approximate
decomposition into largely distinct control knobs: diversity-preserving (QD) selection
sustains a cycling strategy ecology that truncation collapses, while tested selection
variants do not improve realism; reflexive
price feedback raises stylized-fact realism with selection held fixed; amplified
behavioral bias raises a fragility proxy without touching realism; and consensus
network topology moves nothing robustly. Selection is the diversity (and cycling)
knob, microstructure is the realism knob, behavior is the fragility knob---and not
every plausible mechanism is a lever.

{\small
\bibliographystyle{plainnat}
\bibliography{references}
}
\end{document}